\documentclass{article}

\usepackage{PRIMEarxiv}

\usepackage[utf8]{inputenc} % allow utf-8 input
\usepackage[T1]{fontenc}    % use 8-bit T1 fonts
\usepackage{hyperref}       % hyperlinks
\usepackage{url}            % simple URL typesetting
\usepackage{booktabs}       % professional-quality tables
\usepackage{amsfonts}       % blackboard math symbols
\usepackage{nicefrac}       % compact symbols for 1/2, etc.
\usepackage{microtype}      % microtypography
\usepackage{lipsum}
\usepackage{fancyhdr}       % header
\usepackage{graphicx}       % graphics
\usepackage{comment}
\usepackage{amsmath}% 
\usepackage{subfig}
\graphicspath{{media/}}     % organize your images and other figures under media/ folder

\title{Neuromorphic Correlates of Artificial Consciousness}

\author{
  Anwaar Ulhaq \\
  Central Queensland University,  \\
  School of Engineering and Technology\\
  Sydney campus,  Australia\\
  \texttt{\{a.anwaarulhaq@cqu.edu.au} \\
}

\begin{document}
\maketitle

\begin{abstract}
The concept of neural correlates of consciousness (NCC), which suggests that specific neural activities are linked to conscious experiences, has gained widespread acceptance. This acceptance is based on a wealth of evidence from experimental studies, brain imaging techniques such as fMRI and EEG, and theoretical frameworks like integrated information theory (IIT) within neuroscience and the philosophy of mind. This paper explores the potential for artificial consciousness by merging neuromorphic design and architecture with brain simulations. It proposes the Neuromorphic Correlates of Artificial Consciousness (NCAC) as a theoretical framework. While the debate on artificial consciousness remains contentious due to our incomplete grasp of consciousness, this work may raise eyebrows and invite criticism. Nevertheless, this optimistic and forward-thinking approach is fueled by insights from the Human Brain Project, advancements in brain imaging like EEG and fMRI, and recent strides in AI and computing, including quantum and neuromorphic designs. Additionally, this paper outlines how machine learning can play a role in crafting artificial consciousness, aiming to realise machine consciousness and awareness in the future.
\end{abstract}

% keywords can be removed
\keywords{Neural correlates of consciousness (NCC), Integrated information theory (IIT), Artificial consciousness}

\section{Introduction}
Consciousness, originating from the Latin conscius (con- "together" and scio "to know"), pertains to the awareness of both internal and external existence \cite{lewis1990studies}. The modern concept of consciousness is attributed to John Locke, who defined it as "the perception of what passes in a man's own mind" \cite{locke1847essay}. The most challenging aspect of understanding consciousness is often referred to as the 'hard problem' of qualia. This encompasses the subjective experiences associated with phenomena like the specific quality of redness or the feeling of pain \cite{chalmers1997conscious}. At its core, consciousness is the state or quality of being aware of and able to perceive one's surroundings, thoughts, emotions, and experiences. It is also described as the subjective experience of "feeling like something." Historically, it has been a topic of discussion for philosophers, psychologists, and neuroscientists, but the realm of artificial intelligence has brought it into a new spotlight in search of the notion of artificial consciousness. Given our limited or non-uniform understanding of consciousness, a fundamental question is whether machines could achieve artificial consciousness. 

The philosophy of consciousness has a deep-rooted history, originating from the concept of an immortal soul in Greek philosophy \cite{jorgensen2018consciousness} and Avicenna's synthesis of 'nafs' and rational enquiry, 'aqal' \cite{jamali2019avicenna}. These early ideas set the stage for Descartes' influential mind-body dualism \cite{alanen1989descartes}, which became a cornerstone for further philosophical enquiries. Modern perspectives, such as Immanuel Kant's transcendental idealism \cite{allison2006kant} and the phenomenological approach advocated by thinkers like Husserl and Heidegger \cite{stapleton1984husserl}, delve into the active role of consciousness in shaping our perception and understanding of reality. Collectively, these perspectives weave a multifaceted narrative, enriching our exploration of consciousness's nature and its profound impact on human existence and philosophical enquiry.

\begin{figure}
  \centering
  \includegraphics[width=0.8\textwidth]{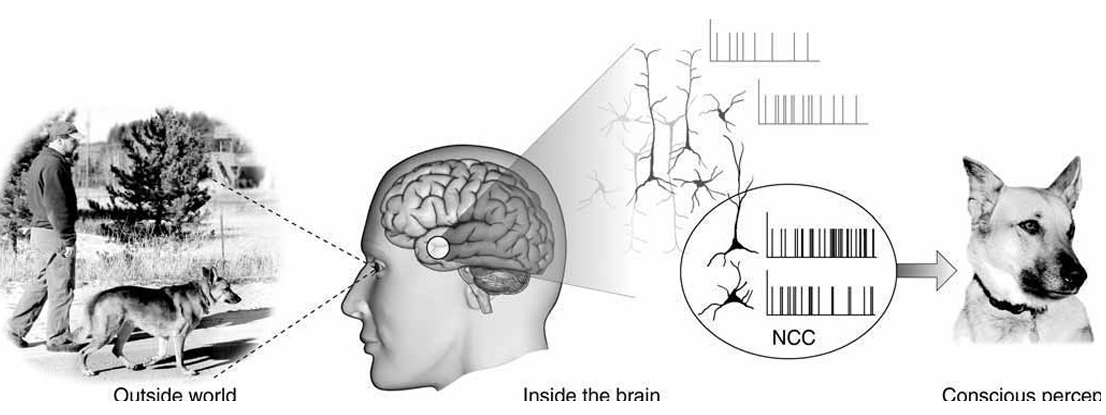}
  \caption{Schema of the neural processes underlying consciousness, from Christof Koch's research, illustrating the intricate interplay of neural correlates of consciousness (NCC) and their dynamic interactions within the brain's complex network. Koch's work sheds light on the fundamental mechanisms that give rise to conscious experience, offering valuable insights into the nature of cognition and subjective awareness. Image source: \cite{koch2004quest}}
  \label{fig:fig1}
\end{figure}

Building upon the rich philosophical foundations, the pursuit of understanding consciousness has been further enriched by groundbreaking developments in neuroscience, particularly the exploration of neural correlates of consciousness (NCC) \cite{koch2004quest}. The advent of advanced brain imaging techniques, such as functional magnetic resonance imaging (fMRI) and electroencephalography (EEG), has allowed scientists to peer into the neural correlates of conscious experience. Pioneering theories like the global workspace theory, proposed by Bernard Baars \cite{baars2005global}, and the integrated information theory, developed by Giulio Tononi \cite{tononi2004information}, have sought to map the neurobiological underpinnings of conscious awareness. Moreover, the discovery of specific brain regions and networks, such as the default mode network \cite{raichle2015brain} and other supporting networks \cite{demertzi2013consciousness}, has shed light on the intricate interplay between conscious and unconscious processes. 

The study of consciousness has entered a new phase with the rise of artificial intelligence (AI) and the quest for artificial general intelligence (AGI). This development has sparked a compelling debate about whether machines can exhibit a form of artificial consciousness comparable to human cognition. As researchers delve deeper into creating intelligent systems that mimic human-like reasoning and understanding, exploring the potential for artificial consciousness has become a captivating frontier in scientific enquiry, igniting discussions and reflections within the scientific community.

The thought experiments of the "philosophical zombie" and "fading qualia" are neutral in their stance regarding the prospect of artificial consciousness \cite{chalmers1997conscious}. Instead, they serve as tools for exploring the nature of consciousness itself, regardless of whether it arises in biological or artificial systems. The philosophical zombie scenario questions whether a being that behaves identically to a conscious being but lacks subjective experience can truly be considered conscious. This highlights the intrinsic difficulty in defining and quantifying subjective awareness, a core aspect of the hard problem. Similarly, fading qualia scenarios explore the gradual reduction or alteration of subjective experiences while maintaining normal external behaviours. These experiments raise fundamental questions about subjective experience, awareness, and the relationship between consciousness and physical processes, challenging our understanding of what it means to be conscious. In the context of artificial consciousness, these experiments highlight the complexity and depth of the challenges involved without explicitly advocating for or against the possibility of achieving genuine artificial consciousness.

\begin{figure}
  \centering
  \includegraphics[width=0.8\textwidth]{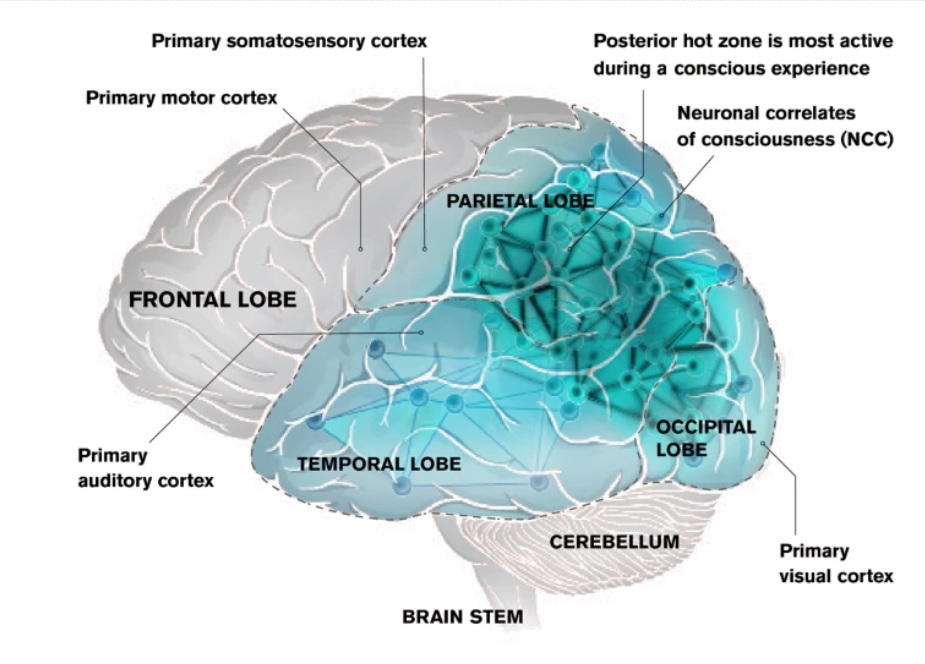}
  \caption{According to Koch, each conscious experience is associated with a specific integrated state in the brain's posterior hot zone. This hot zone encompasses the parietal, occipital, and temporal lobes of the cerebral cortex. Image Source: \cite{koch2018consciousness}}
  \label{fig:fig2}
\end{figure}

Building on philosophical and neuroscientific foundations, this paper introduces a novel framework for exploring artificial consciousness. We optimistically argue that integrating neuromorphic design and brain simulations and proposing \textbf{neuromorphic correlates of artificial consciousness (NCAC)} allow for the systematic investigation into artificial consciousness's feasibility. As neural understanding deepens, replicating these processes in artificial systems may become feasible, supported by initiatives like the BRAIN Initiative and projects like the Human Brain Project and the Blue Brain Project. While contentious, our work invites critical discourse to advance the understanding of consciousness and its potential in artificial systems.

The motivation for this work arises from the desire to comprehend and potentially emulate consciousness within artificial systems. Despite the inherent complexities in understanding consciousness, this paper suggests that a theoretical framework can be a foundational tool for envisioning artificial consciousness. The paper establishes the initial framework for research into artificial consciousness by integrating neuromorphic design, brain simulations, and integrated information theory (IIT). The primary contributions of the paper are outlined as follows:

\begin{itemize}
  \item Proposing NCAC as a Theoretical Framework: The paper introduces the concept of neuromorphic correlates of artificial consciousness (NCAC), providing a theoretical basis for studying and understanding artificial consciousness within the context of neuromorphic architectures and brain-inspired simulations.
  
 \item This paper elucidates the components of the proposed framework using an approach that integrates historical developments, established ideas in neuroscience, theories of consciousness, and advancements in computing, AI, brain imaging technologies, and modelling.
\end{itemize}

The paper is structured to provide a comprehensive understanding of artificial consciousness. The first section, "Background Work," covers crucial topics like neural correlates of consciousness, integrated information theory (IIT), brain simulations, and neuromorphic designs. Following this, "Theoretical Basis" delves into the theoretical framework of neuromorphic correlates of artificial consciousness (NCAC). Lastly, "Rationale and Implications" discusses the necessity of artificial consciousness and its potential impact, concluding with a concise summary and references.

\section{Background Work}
In this section, the paper will delve into foundational concepts essential for understanding artificial consciousness. This includes an exploration of the neural correlates of consciousness, which are the specific neural mechanisms or processes associated with conscious experience. Integrated Information Theory (IIT) will be discussed, highlighting its role in providing a theoretical framework for quantifying and understanding consciousness.

\subsection{Neural Correlates of Consciousness: Supporting Evidence and Insights}

Neural Correlates of Consciousness (NCC) is a concept proposed in the field of neuroscience that refers to the minimal neuronal mechanisms or patterns of activity in the brain that are directly associated with or give rise to subjective conscious experiences. Francis Crick and Christof Koch are considered early proponents of the NCC concept, even though they didn't coin the exact term. Their 1990 paper, "Towards a Neurobiological Theory of Consciousness," is a landmark contribution to the field. The term "Neural Correlates of Consciousness" was coined by David Chalmers in his book "The Conscious Mind", published in 1996 \cite{chalmers1997conscious}.

According to Christof Koch \cite{koch2018consciousness}, defining NCC as "minimal" is crucial. Take the cerebellum, a region rich in neurons but not central to subjective experience. Damage to the cerebellum, even substantial, has minimal impact on consciousness. Similarly, injuries to the spinal cord can cause paralysis and loss of sensations, yet conscious experience remains largely intact. This highlights that consciousness isn't solely triggered by neural tissue activity but involves more intricate mechanisms. In a recent article, Christof Koch has proposed footprints of consciousness related to the back of the brain, specifically highlighting regions that play a crucial role in conscious experiences.

\begin{figure}
  \centering
  \includegraphics[width=0.95\textwidth]{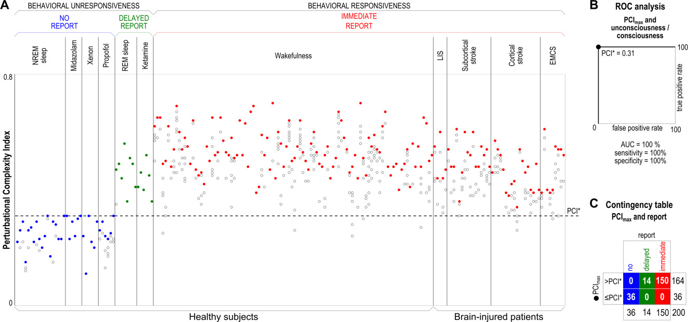}
  \caption{(A) The circles represent the Perturbational Complexity Index (PCI) values derived from cortical responses to transcranial magnetic stimulation (TMS) across various brain stimulation sites. These values are sorted by age and condition, reflecting different states of consciousness, such as non-REM sleep, anesthesia, dreaming, and wakefulness. The solid circles denote the highest PCI value (PCImax) for each individual, while open circles represent lower PCI values. (B) The Area Under the Curve (\text{AUC}) is 100\%, indicating excellent discriminatory power. Using PCI as the cutoff, both sensitivity and specificity achieve 100\%, demonstrating the accuracy of this cutoff in identifying conscious states. (C) The contingency table is created by dividing the PCImax values using the PCI$^*$ cutoff, as illustrated by the dashed horizontal line in panel A. This table categorizes different states, such as emergence from a minimally conscious state (EMCS), locked-in syndrome (LIS), and rapid eye movement (REM), providing a structured view of how the PCI values correlate with various levels of consciousness and brain activity. Image source. \cite{casarotto2016stratification}
}
  \label{fig:fig3}
\end{figure}

The "zap and zip" methodology \cite{koch2017make} is a means to investigate consciousness. This approach involved the application of a sheathed coil of wire to the scalp, delivering an intense magnetic pulse ("zapping") to trigger transient electrical currents in underlying neurones. These neural responses were captured using an array of electroencephalogram (EEG) sensors positioned externally on the skull, generating a real-time depiction of electrical patterns associated with distinct brain regions known as NCC. The "zap and zip" technique estimated the brain's response complexity. Awake volunteers exhibited a "perturbational complexity index" ranging from 0.31 to 0.70, which decreased to below 0.31 during deep sleep or under anaesthesia. The Perturbational Complexity Index (PCI) \cite{casarotto2016stratification} quantifies the complexity of brain responses to stimuli, with higher values associated with wakefulness and conscious awareness. In the context of Neural Correlates of Consciousness (NCC), PCI provides insights into the dynamics of neural activity during conscious states, offering a potential indicator of conscious awareness based on the complexity of brain responses. Recent experiments conducted by Silvia Casarotto et al. \cite{casarotto2016stratification}. corroborate this In a recent study on measuring the Origins of Perinatal Experience by Joel Frohlich and team \cite{frohlich2023not}, they suggest that data derived from MEG, EEG, and fMRI recordings can be utilized to examine theories of consciousness, similar to how cosmologists rely on data from the cosmic microwave background to assess theories about the universe's inception, even though this data is not directly observable or accessible \cite{evans2015cosmic}.

\begin{figure}%
    \centering
    \subfloat[\centering label 1]{{\includegraphics[width=8cm]{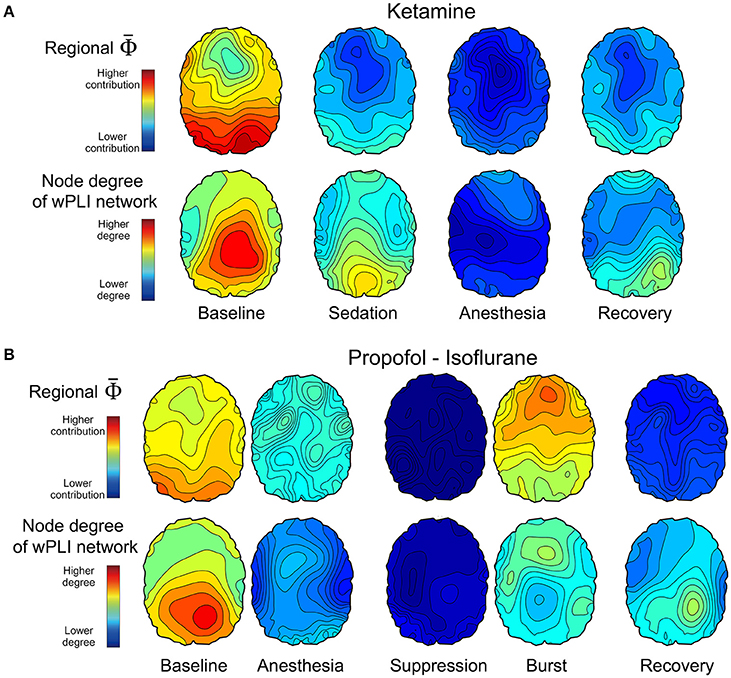} }}%
    \qquad
    \subfloat[\centering label 2]{{\includegraphics[width=6cm]{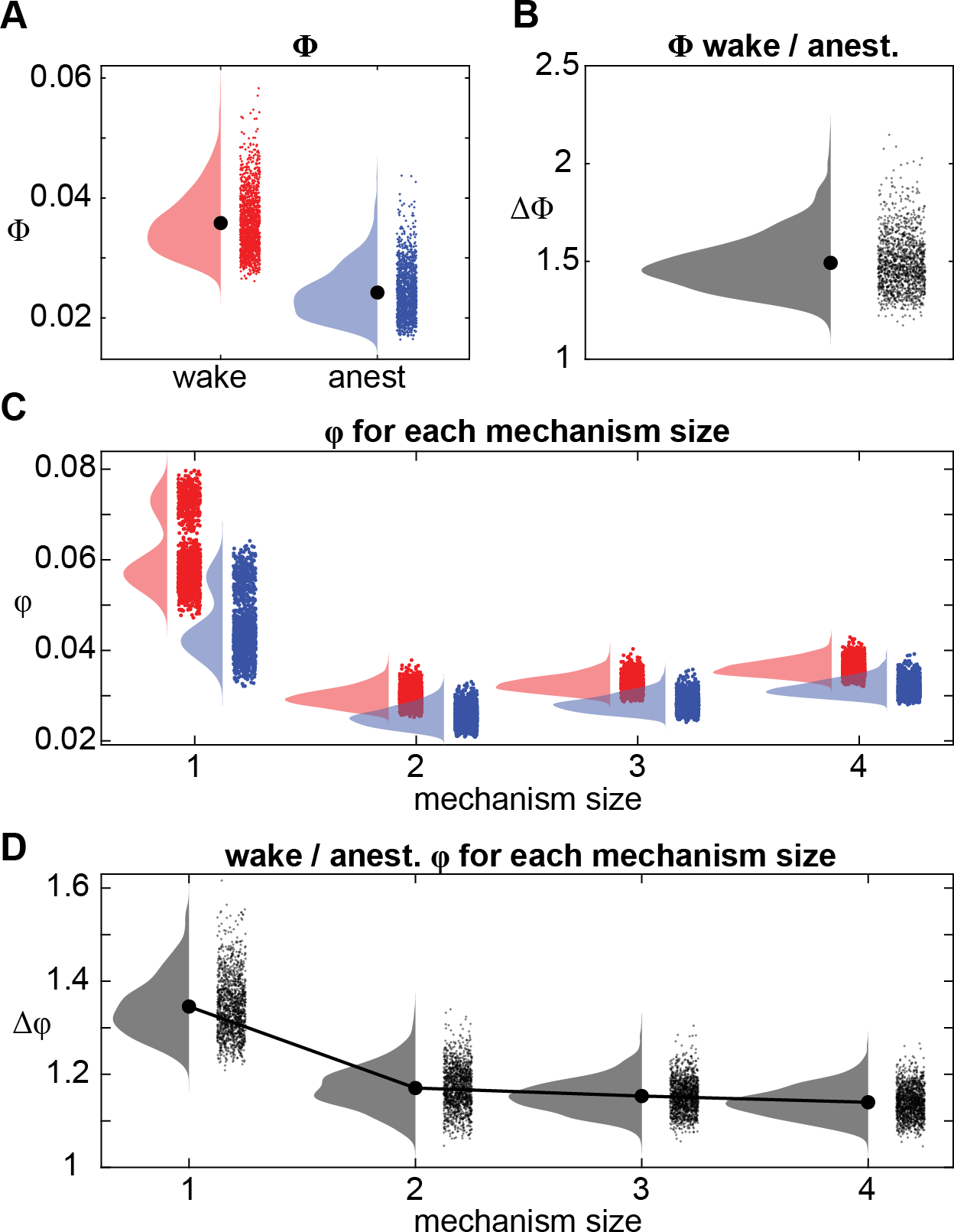} }}%
    \caption{Left: The topographic structures of $\overline{\Phi}_R$ and EEG connectivity in the alpha band were analyzed across states of consciousness during two anesthetic experiments: (A) ketamine and (B) propofol followed by isoflurane. In both experiments, the first row shows the topography structure of $\overline{\Phi}_R$, while the second row presents the node degree of 96 EEG channels, averaged across subjects and states. These structures and strengths reflect the varying levels of consciousness. For instance, higher $\overline{\Phi}_R$ and node degree in the posterior region during the baseline state is disrupted during anesthesia but partly restored upon regaining responsiveness. While the scales of $\overline{\Phi}_R$ and node degree are consistent within each experiment, they differ between the ketamine and propofol-isoflurane experiments. image source \cite{kim2018estimating} Right: The impact of anesthesia on system-level integrated information ($F$) and the integrated information structure (IIS: a set of $\phi$ values) is examined in relation to the fruit fly Drosophila melanogaster. (A) $F$ values are compared between wakefulness (red) and anesthesia (blue) across 1365 channel sets, averaged across flies. (B) The ratio of $F$ (wakeful/anesthetized) is shown for all channel sets, averaged across flies. (C) $\phi$ values from the IIS are compared between wakefulness (red) and anesthesia (blue) for different mechanism sizes, averaged across flies and channel sets. (D) The wakeful $\phi$ to anesthetized $\phi$ ratio is presented for each mechanism size, averaged across flies. Image Source \cite{leung2021integrated} }%
    \label{fig:fig4}
\end{figure}

However, delving into the concept of NCC reveals its complexity. While alterations in brain activity often align with shifts in consciousness, establishing direct causation isn't straightforward; external factors may also contribute. Even with a comprehensive map of the NCC, unravelling the mechanisms by which brain processes give rise to subjective experiences (the "hard problem" of consciousness) remains an elusive endeavour. Yet, ongoing research continuously fortifies the link between specific brain patterns and conscious states, underscoring the importance of this exploration in understanding the fundamental nature of consciousness.

\subsection{Integrated Information Theory (IIT): Supporting Evidence and Insights}

The Global Neuronal Workspace (GNW) theory \cite{baars2005global} suggests that consciousness emerges when sensory data are globally broadcasted to different cognitive systems for processing, storage, retrieval, or action execution, akin to a "dashboard" in early artificial intelligence systems. This global broadcasting, constrained by capacity, permits only a portion of information to be consciously processed, resulting in subjective awareness. However, in science, we primarily deal with numbers and quantifiable data. Integrated Information Theory (IIT) provides a quantitative perspective with a mathematical approach, offering insights into the mechanisms underlying consciousness.

Integrated Information Theory (IIT), proposed by neuroscientist Giulio Tononi \cite{tononi2004information}, offers a unique mathematical approach to understanding consciousness. While not claiming exclusivity, IIT provides a quantifiable framework. Its core axioms include intrinsic existence, emphasising a system's self-contained nature regardless of external interactions, which is essential for grasping consciousness. Composition asserts that a system's structure defines its states and transitions, as illustrated by neural networks' states and connections in the brain. The information axiom highlights causal dependencies in a system's past and current states, which is crucial in understanding how neural activity influences consciousness. Integration shows how systems generate more information than their parts, as exemplified by emergent properties in the brain. Lastly, the exclusion axiom underscores filtering irrelevant information, crucial for efficient cognitive functioning, reflecting IIT's quantitative perspective on consciousness.

\begin{figure}
  \centering
  \includegraphics[width=0.9\textwidth]{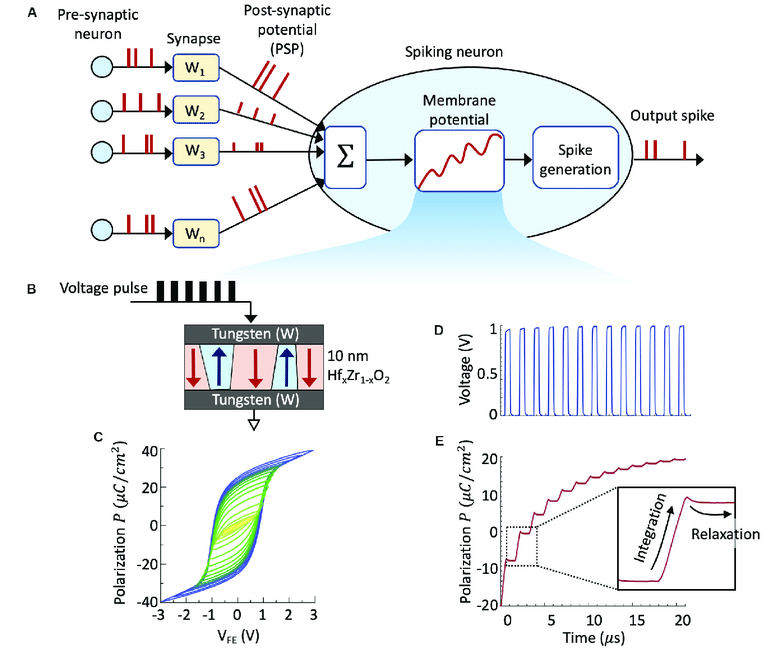}
  \caption{(A) Spiking Neural Network: This model represents a neuron (spiking) connected by adjustable synapses (plastic) in an array. The core of the neuron is its membrane, modelled by a special state variable (ferroelectric polarization). (B) Ferroelectric Capacitor: This tiny device (MFM capacitor) uses a thin film (Hf x Zr 1-x O 2) sandwiched between metal plates (W) to study how voltage affects polarization switching. (C) Polarization vs. Electric Field: This graph shows how the material's polarization (P) changes with applied voltage (E). The loops indicate multiple internal regions within the film. (D-E) Mimicking Neurons: Short voltage pulses can be used to study how polarization changes over time. This behaviour resembles how a real neuron's membrane integrates and relaxes with electrical signals. Image Source \cite{dutta2020supervised}}
  \label{fig:fig5}
\end{figure}

To quantify integrated information, IIT introduces the concept of phi ($\Phi$), which measures the amount of integrated information within a system. The calculation of phi involves several steps:

\begin{enumerate}
  \item \textbf{Partitioning:} Divide the system into non-overlapping parts or subsets, considering all possible partitions.
  \item \textbf{Cause-Effect Repertoire:} Determine the cause-effect repertoire for each partition, which specifies how each subset's states depend on past states within the subset and across partitions.
  \item \textbf{Information Quantification:} Calculate the information generated by each partition and the information generated by the system as a whole.
  \item \textbf{Integration Measure:} Compute phi ($\Phi$) as the difference between the information generated by the system as a whole and the sum of information generated by its parts, normalized to account for system size.
\end{enumerate}

The formula for calculating phi is:

\[
\Phi = \Phi^\text{max} = \max \left( \sum_{\text{partition}} \Phi_\text{part}, 0 \right)
\]

Here, $\Phi^\text{max}$ represents the maximum integrated information across all possible partitions, and $\Phi_\text{part}$ represents the integrated information for each partition.

Calculating Phi ($\Phi$), or Integrated Information, in Integrated Information Theory (IIT) presents a significant challenge due to its inherent complexity. Here's a breakdown of why it's so time-consuming:

  Imagine a system with just 10 elements, each with two possible states (on/off). The number of possible system states becomes $2^{10}$ (1024). To calculate $\Phi$, IIT needs to analyse the causal interactions between all these states, considering how a change in one element affects the possibilities of others. This complexity increases exponentially with the number of elements and states. In a real brain with billions of neurons, the number of states becomes astronomically large, making a complete calculation practically impossible. Even if Phi could be accurately calculated, there's no established "consciousness meter" to validate its meaning. While Phi isn't yet a measure of consciousness, IIT provides a valuable framework for understanding the potential role of information integration in conscious experience.  

A recent work by Hyoungkyu Kim introduced an innovative and practical approach for estimating $\Phi$ from high-density EEG data. This method was then applied to examine different states of consciousness altered by general anaesthesia induced by ketamine and propofol-isoflurane. By investigating the EEG properties associated with large and small $\overline{\Phi}$ values, the study inferred the large-scale network correlates of various consciousness states. The utilisation of a multi-dimensional parameter space, which included various EEG-derived measures of connectivity along with $\overline{\Phi}R$, proved effective in distinguishing between different states of consciousness and sub-states during burst and suppression phases. The integration of $\Phi$ with EEG connectivity within clinically defined anaesthetic states presents a novel practical approach to applying IIT. This method has the potential to characterise a range of physiological (sleep), pharmacological (anaesthesia), and pathological (coma) states of consciousness in the human brain.

\begin{figure}%
    \centering
    \subfloat[\centering label 1]{{\includegraphics[width=3.5cm]{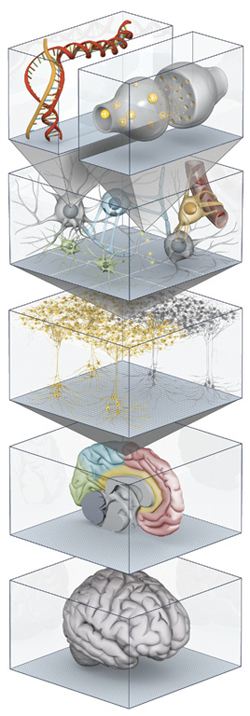} }}%
    \qquad
    \subfloat[\centering label 2]{{\includegraphics[width=10cm]{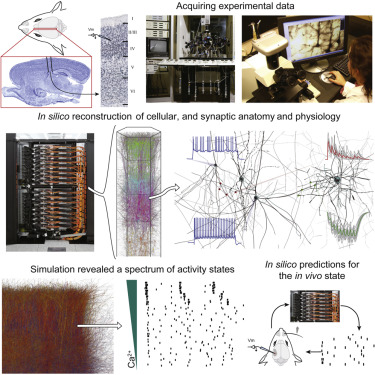} }}%
    \caption{Left: The abstract of the Human Brain Project to simulate various levels of brain function, spanning from chemical and electrical signalling to the cognitive traits underlying intelligent behaviours, encompassing Molecular, Cellular, Circuits, Regions, and the Whole Organ. Image source \cite{markram2012human} Right: The Blue Brain Project focused on simulating cortical columns of the mouse brain. Image source \cite{markram2015reconstruction}}%
    \label{fig:fig7}%
\end{figure}

Another group  \cite{leung2021integrated} studied the information structure of the fruit fly's (Drosophila melanogaster) brain in both wakefulness and under general anaesthesia. They found that when the flies were awake, there were complex and interconnected patterns of information exchange among groups of neurones. However, when the flies were under anaesthesia, these complex patterns collapsed into isolated clusters of interactions. The study's findings align with the core idea of IIT, which proposes that consciousness is linked to the integrated processing of information within a system. The observed collapse of integrated interactions among neurones during anaesthesia and the presence of complex information structures during wakefulness is in line with IIT's predictions about the nature of consciousness. Overall, this experimental support for IIT and the interpretation of phi in the context of consciousness has a significant impact on trusting IIT as a valuable and insightful theory in the field of neuroscience and consciousness studies.

\begin{figure}
  \centering
  \includegraphics[width=0.65\textwidth]{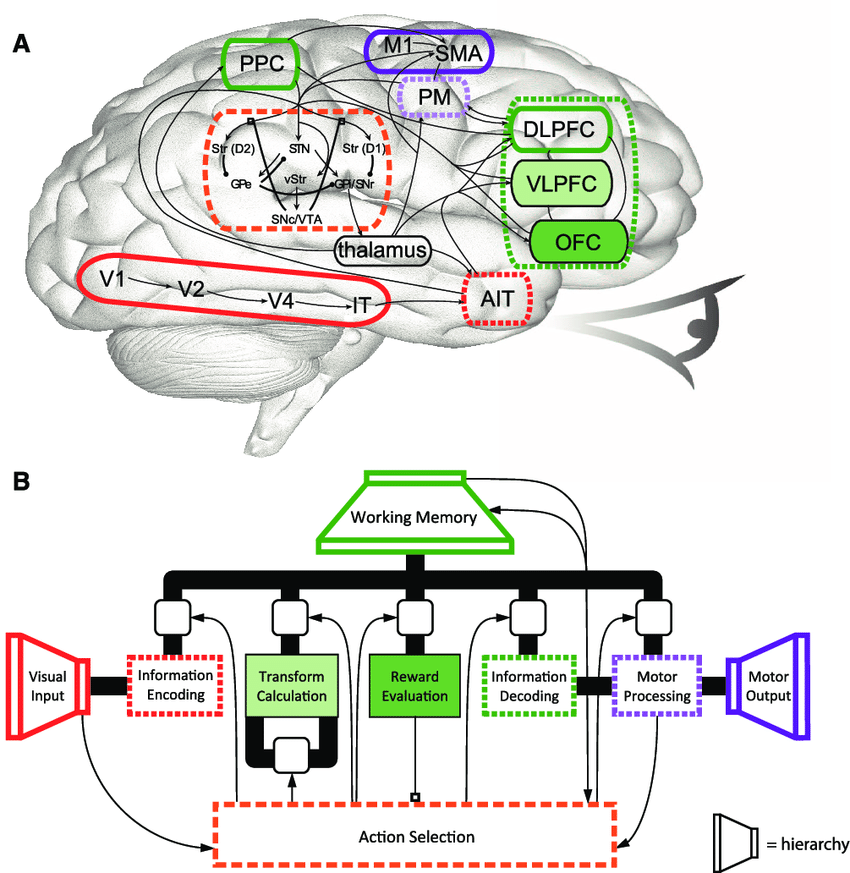}
  \caption{This illustration represents a large-scale 3D brain model, inspired by Spaun.(a) It highlights the main brain structures of Spaun and their connections. (b) It depicts communication pathways within Spaun's brain: thick black lines represent connections within the cortex, while thin lines show connections between the Basal Ganglia and the cortex. Image source: \cite{vu2019fault} }
  \label{fig:fig6}
\end{figure}

\subsection{Spiking Neural Networks (SNNs) and Neuromorphic Computing: Progress and Insights}

Neuromorphic computing emulates the brain's structure and function by utilising artificial neurones and connections akin to biological brains, aiming for brain-like processing capabilities. Spiking neural networks (SNNs) within neuromorphic computing mimic biological networks, processing information via sparse binary spikes representing neuronal firing times. These spikes reduce computational load and energy usage compared to continuous processing in traditional Artificial Neural Networks (ANNs), as only active neurons contribute to computations due to the spiking dynamics and network sparsity. Neurons in SNNs integrate incoming spikes using mechanisms like leaky integrate-and-fire models, firing spikes based on threshold criteria, and simplifying computations to basic additions and subtractions.

The Linear-Leaky-Integrate-and-Fire (LLIF) neuron model \cite{lu2022linear}, used in neuroscience and neural network simulations, mimics biological neurons' input integration with a leaky membrane, gradually returning the neuron's potential to rest. This leakage, controlled by the membrane time constant, results from ion flow through the membrane, restoring the resting potential after activity. Inputs are linearly combined, triggering a spike if they surpass a set threshold, followed by a reset akin to biological refractory periods. In Spiking Neural Networks (SNNs), Spike-Timing-Dependent plasticity (STDP) \cite{caporale2008spike} is vital for unsupervised learning, adjusting synaptic weights based on spike timing. STDP enables SNNs to learn patterns from input spike sequences without labelled data, strengthening connections between co-activated neurones through mechanisms like Hebbian learning and forming functional circuits encoding input features.

Traditional artificial neural networks (ANNs) use backpropagation to compute gradients for optimising model parameters, but applying backpropagation directly to spiking neural networks (SNNs) is complex due to its event-driven and spike-based nature. Instead, SNNs utilise surrogate gradient descent methods, employing surrogate objectives or learning rules easier to compute gradients for, such as spike-timing-dependent plasticity (STDP) rules or firing rates. While SNNs can simulate neural activity, capturing neural correlations of consciousness (NCC) in a comprehensive manner remains challenging, as subjective aspects of consciousness are difficult to represent computationally. NCC likely involves higher-level cognitive processes and global information integration, which may surpass the capabilities of traditional SNN models, highlighting the limitations of emulating consciousness with current computational approaches.

As of now, Spiking Neural Networks (SNNs) can model networks with thousands to tens of thousands of neurones efficiently. However, scaling SNNs to larger models with hundreds of thousands or millions of neurons can be challenging due to computational and memory constraints. The current capacity of SNNs is primarily limited by hardware capabilities, such as processing power and memory bandwidth, as well as the complexity of simulating spike dynamics accurately in large-scale networks while maintaining real-time performance. Advances in hardware and algorithm optimisation are ongoing to overcome these limitations and enable future simulation of even larger SNNs.

\subsection{Neuron Dynamics and Brain Simulations: Projects and Insights}

Brain simulations offer a unique pathway to understanding brain dynamics, neuron behaviour, and, ultimately, consciousness. By replicating the intricate interactions among neurones and neural networks, simulations provide a controlled environment for studying complex neural processes. This approach allows researchers to observe how individual neurones communicate, how networks form, and how information flows within the brain. Such simulations can model various brain states and responses to stimuli and even simulate pathological conditions, providing insights into neurological disorders. Ultimately, by bridging the gap between theory and experimentation, brain simulations offer a powerful tool for advancing our understanding of brain function and the mechanisms underlying consciousness.

Brain simulations like The Blue Brain Project and the Human Brain Project (HBP) ran on supercomputers or high-performance computing clusters, utilizing specialised software frameworks and algorithms tailored for brain modelling. The HBP also investigated neuromorphic computing, leveraging hardware inspired by brain architecture for simulation and computation. However, technological constraints limited the ability to achieve a complete simulation of a human brain.

The Blue Brain Project developed detailed digital replicas of the mouse brain, exploring cellular and synaptic complexities. A notable milestone was simulating a segment of the rat cortex, involving 4 million neurons and 14 billion connections. This achievement showcased high-resolution brain simulations, yet it paled in comparison to the vast scale of the human brain's billions of neurons and trillions of synaptic connections. Current supercomputers lack the processing power for such a scale, but future advancements may enable larger and more complex brain simulations.

A more adaptable approach involves using a virtual model to replicate brain functionality. The Semantic Pointer Architecture Unified Network (Spaun) is an artificial model of the human brain developed by Professor Chris Eliasmith and colleagues at the University of Waterloo in Canada. It comprises approximately two and a half million virtual neurones organised into functional groups, mirroring the arrangement of real neurones in various brain regions responsible for functions like vision and short-term memory. (For context, the human brain has about 100 billion neurons.)

Spaun  \cite{vu2019fault} is designed to process sequences of visual images across eight distinct tasks. It analyses the presented information and makes decisions based on this analysis. For instance, Spaun can recognise and memorise numbers written in different styles, and it can replicate them using a mechanical arm. Moreover, Spaun demonstrates the ability to answer numerical questions and complete number series after being provided with examples, showcasing its versatile cognitive capabilities.

However, achieving a truly realistic artificial consciousness remains a formidable challenge. The human brain's complexity, with billions of neurones and trillions of synaptic connections, is currently beyond the reach of even the most powerful supercomputers. Additionally, understanding consciousness involves more than just replicating cognitive functions; it requires grasping subjective experiences, emotions, self-awareness, and other higher-order aspects of consciousness that are not fully understood.

\section{Theoretical Basis for Neuromorphic Correlates of Artificial Consciousness (NCAC)}

In Integrated Information Theory (IIT), the concept of Neural Correlates of Consciousness (NCC) pertains to the specific neural mechanisms or processes that are associated with consciousness. The Phi ($\Phi$) value in IIT represents the level of integrated information within a system, indicating its capacity for generating complex causal relationships and irreducible wholes, which are considered crucial for consciousness.

If Spiking Neural Networks (SNNs) can successfully emulate Neural Correlates of Consciousness (NCC) in a manner convincingly replicating aspects of conscious experiences, it could be termed the Development of Neuromorphic Correlates of Artificial Consciousness. This stage would denote the minimal artificial neural circuits, parameters or processes within a neuromorphic system corresponding to artificial manifestations of consciousness. The concept of "Development of Neuromorphic Correlates of Artificial Consciousness" encapsulates the idea that specific neural patterns or dynamics learned by SNNs through machine learning and implemented in neuromorphic hardware can lead to artificial states of consciousness in machines. 

From an Integrated Information Theory (IIT) perspective, Neuromorphic Correlates of Artificial Consciousness (NCACs) trained to maximise high $\Phi$ values are desirable as they develop artificial systems capable of complex and integrated information processing similar to conscious systems.

Now, let us propose a theoretical framework using a hypothetical process for potentially achieving artificial consciousness through spiking neural networks (SNNs) based on principles from integrated information. 

This framework is divided into four phases.

\textbf{Quantification}: For any digital system, quantification is key. At this juncture, it is imperative to substantiate the empirical linkage between neural correlates of consciousness (NCC) and the various stages of consciousness observed in both humans and other primates. Existing empirical evidence strongly supports the notion that NCC is an outcome of consciousness rather than its antecedent. This association constitutes a critical foundation for the progression of subsequent phases within the framework. Emerging technologies such as fMRI, EEG, and TMS could be instrumental at this stage, particularly when combined with subjective evaluations of human subjects. The association between NCC and consciousness must be quantified to validate consciousness and ascertain its presence. An existing quantification framework, integral to integrated information theory (IIT), utilises high values of $\Phi$ to denote heightened levels of consciousness and, conversely, low values to indicate diminished consciousness. While empirical findings from various experiments lend support to this theory, generalisation remains challenging at this stage. Nevertheless, a validated association through quantification will facilitate the integration of artificial consciousness into machines.

\textbf{Simulation}: A critical stride toward achieving artificial consciousness involves simulating and emulating neural connectivity, dynamics, and patterns using brain-inspired architectures. This approach encompasses projects such as the Blue Brain Project, Human Brain Project, or EBRAINS, as well as entirely virtual simulations of brain functionality like Spaun. The objective behind these simulations is to replicate neural circuits and functionalities observed in neuroanatomy with functional connectivity. Brain-inspired systems, such as Spiking Neural Networks, with their intricate architectures and replication of asynchronous and synchronous brain activity, play a pivotal role in emulating these behaviours. The simulation should ideally be energy-efficient and possess time and space complexities close to those of the brain, enabling the firing, activation, or deactivation of desired circuits akin to artificial anaesthesia and wake-up processes in living organisms. While there are examples of projects demonstrating reasonable success in this endeavour, current technology and computing capabilities still lag far behind in simulating the complete circuitry necessary for artificial consciousness.

\textbf{Adaptation}: In the adaptation phase, machine learning assumes critical importance as the simulated brain model undergoes iterative refinement to replicate the intricacies of conscious experience faithfully. The emphasis is placed on capturing neural connectivity and activity patterns pivotal for emulating artificial consciousness. Unlike traditional datasets in machine learning, these datasets include brain connectivity patterns or specifications of neural correlates of consciousness (NCCs), leading to various stages of consciousness in living brains. Consequently, the simulated brain can be trained using mathematical optimisation techniques to reduce dissimilarities or minimise loss in a supervised fashion. For instance, a loss function can then be defined as the difference between the $\Phi$ values of NNACs and the $\Phi$ values of known NCCs, representing the gap between artificial and biological correlates of consciousness. Inspired by transfer learning principles, the model assimilates these extensive datasets, enabling it to encode and understand the nuanced associations between specific neural firing patterns and conscious experiences. Continuous feedback loops, akin to reinforcement learning, guide the model's optimisation, directing the firing of specific neurones to emulate the dynamic nature of consciousness. Fine-tuning mechanisms allow for meticulous adjustments, ensuring alignment with empirical observations of conscious phenomena. By integrating novel insights from neuroscience, the adapted simulated models progressively converge towards replicating the complexities of conscious experience, thereby facilitating the transfer of conscious-like functionalities to machines. This phase is pivotal for establishing associations between NCC and artificial consciousness, bridging the gap between neural correlates and simulated conscious states.

\textbf{Implementation}: The final stage entails implementing neuromorphic architectures in hardware to facilitate their deployment in physical systems, such as robots, or the implantation of neuromorphic chips into human subjects lacking conscious experiences due to any disorder. A major challenge in this phase revolves around qualia, which concerns subjective notions of experience. However, this obstacle could be mitigated by integrating an expressive artificial intelligence infrastructure into machines. Such infrastructure would enable machines to express experiences akin to humans, allowing for the verification of qualia, such as sharing the sensation of redness or feelings of pain. Additionally, ensuring seamless integration of neuromorphic hardware with existing systems and environments presents another significant hurdle. Robust testing protocols and rigours ethical considerations within an established ethical framework will be essential throughout the implementation process, particularly when deploying neuromorphic technologies in human subjects. Despite these challenges, the potential benefits of neuromorphic architectures in enhancing human-machine interactions and restoring consciousness in individuals with disorders underscore the importance of continued research and development in this field.

\begin{figure}
  \centering
  \includegraphics[width=0.6\textwidth]{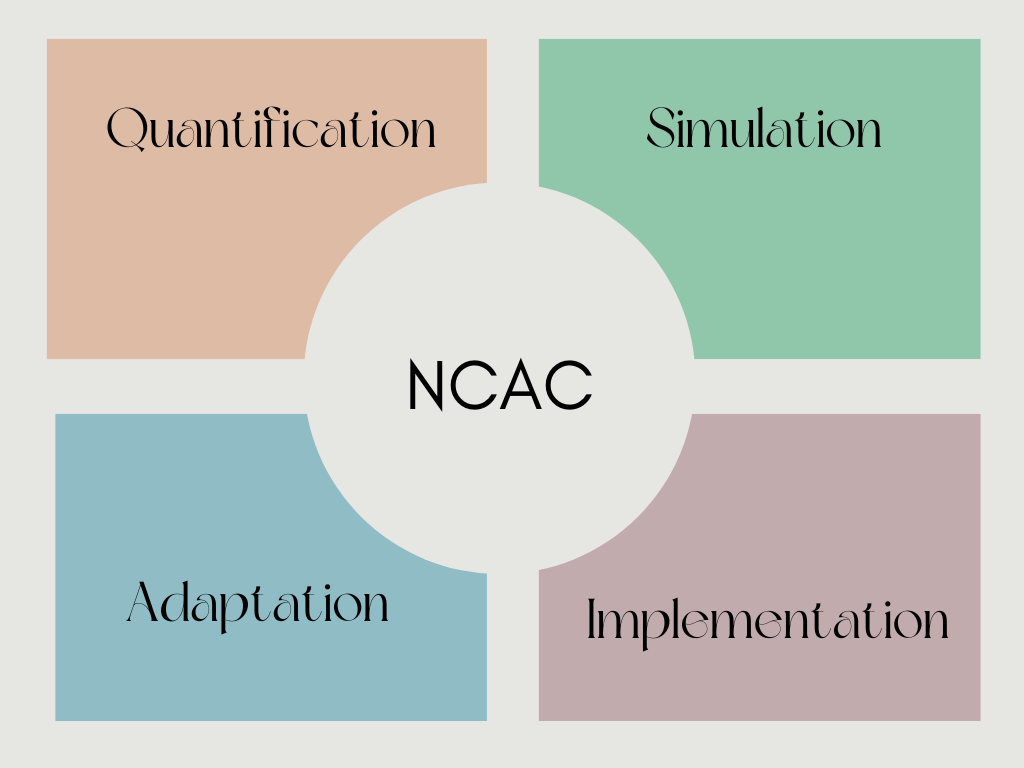}
  \caption{Theoretical Framework: NCAC (Neuromorphic Correlates of Artificial Consciousness), proposing artificial consciousness through neuromorphic design aligned with integrated information principles. The framework unfolds across four phases: Quantification, Simulation, Adaptation, and Implementation.}
  \label{fig:fig5}
\end{figure}

\subsection{Assumptions}

\begin{itemize}
    \item \textbf{Correlation Between NCC and High $\Phi$ Values}: A significant correlation exists between neural correlates of consciousness (NCC) in biological systems and the presence of a consciousness metric, such as $\Phi$ values, as defined by Integrated Information Theory (IIT).
    
    \item \textbf{Feasibility of NNAC Simulation}: Simulations of Neuromorphic Correlates of Artificial Consciousness (NNAC) using Spiking Neural Networks (SNNs) or similar neuromorphic architectures are feasible, enabling the replication of neural mechanisms, connectivity, and dynamics identified in NCC.
    
    \item \textbf{Optimization}: Implementing an optimisation method to minimize the disparity between NNACs' and NCCs' $\Phi$ values effectively bridges the gap between artificial and biological correlates of consciousness. For example, employing a gradient descent-like algorithm can efficiently optimize NNACs by adjusting the parameters (weights) of the SNNs or neuromorphic models to minimise the loss function.
    
    \item \textbf{Convergence of Optimisation}: The optimisation process will converge to a solution where the $\Phi$ values of NNACs closely resemble those of NCCs, indicating a high level of similarity in terms of integrated information and the potential for artificial consciousness. Moreover, domain transfer becomes feasible as well.
\end{itemize}

\section{Rationale and Implications of Artificial Consciousness: A Discussion}
The motivation behind developing artificial consciousness stems from a variety of interconnected factors that span the scientific, technological, ethical, and philosophical realms. At its core, the pursuit of artificial consciousness is driven by a fundamental curiosity about the nature of consciousness itself. By attempting to replicate aspects of consciousness in machines, researchers aim to unravel the mysteries of subjective experience, self-awareness, and intentionality. This exploration not only deepens our understanding of the mind but also contributes to the advancement of artificial intelligence (AI). Artificially conscious systems represent a significant milestone in AI development, going beyond mere algorithmic processing to embody qualities such as empathy, context awareness, and adaptive learning.

Ethical considerations also play a crucial role in the quest for artificial consciousness. As AI systems become increasingly integrated into society, questions of accountability, transparency, and fairness arise. Developing artificial consciousness with ethical principles in mind can lead to responsible AI that aligns with human values and respects individual rights. Moreover, the potential applications of artificially conscious systems are vast and diverse. From enhancing human-machine interactions in areas like human-computer interfaces, robotics, and virtual assistants to revolutionising healthcare with empathetic AI, the impact of artificial consciousness spans across various domains.

Ultimately, the pursuit of artificial consciousness represents a multidisciplinary endeavour that bridges disciplines such as neuroscience, computer science, philosophy, and psychology. It embodies a quest for innovation, understanding, and ethical advancement in AI and human-machine interactions.

\section{Conclusion} %%%%%%%%%%%%%%%%%%%%%%%%%%%%%%%%%%%%%%%%
\label{S-Conclusion} 
In conclusion, this paper explored the potential for artificial consciousness through the integration of neuromorphic design and architecture with brain simulations, proposing the Neuromorphic Correlates of Artificial Consciousness (NCAC) as a theoretical framework. While the debate on artificial consciousness has been contentious due to our incomplete understanding of consciousness, this work aims to contribute to the ongoing dialogue in the field. The optimistic and forward-thinking approach taken in this study was fueled by insights from the Human Brain Project, advancements in brain imaging such as EEG and fMRI, and recent progress in AI and computing, including quantum and neuromorphic designs. Furthermore, the paper outlined the potential role of machine learning in shaping artificial consciousness, with the aim of realising machine consciousness and awareness in the future.

%%%%%%%%%%%%%%%%%%%%%%%%%%%%%%%%%%%%%%%%%%%%%%%%%%%%%%%%%%%%%%%%%%%%%%%%%%%

%Bibliography
%\bibliographystyle{unsrt}  
%\bibliography{references}  

\end{document}